\documentclass[conference]{IEEEtran}
\IEEEoverridecommandlockouts
\usepackage{cite}
\usepackage{amsmath,amssymb,amsfonts}
\usepackage{algorithmic}
\usepackage{graphicx}
\usepackage{caption}
\usepackage{subcaption}
\usepackage{textcomp}
\newcommand{\R}{\mathbb{R}}

\DeclareMathOperator*{\argmin}{arg\,min}
\usepackage{xcolor}
\def\BibTeX{{\rm B\kern-.05em{\sc i\kern-.025em b}\kern-.08em
    T\kern-.1667em\lower.7ex\hbox{E}\kern-.125emX}}
\usepackage{algorithm,algorithmic}
\usepackage{mathtools}
\newtheorem{mythm}{Theorem}

\newtheorem{lemma}[mythm]{Lemma}
\usepackage{url}
\usepackage{hyperref}
\hypersetup{
    colorlinks=true,
    linkcolor=blue,
    filecolor=magenta,      
    urlcolor=cyan,
    pdftitle={Overleaf Example},
    pdfpagemode=FullScreen,
    }
\begin{document}

\title{An Image Segmentation Model with Transformed Total Variation
\thanks{The work was partially supported by NSF grants DMS-2151235, DMS-2219904, and a Qualcomm Faculty Award.}
}
\author{$\textnormal{Elisha Dayag}^1$, $\textnormal{Kevin Bui}^1$, $\textnormal{Fredrick Park}^2$, and $\textnormal{Jack Xin}^1$
\\
$^{1}$Department of Mathematics;
University of California, Irvine;
Irvine, CA 92697, United States \\
$^{2}$Department of Mathematics \& Computer Science;
Whittier College; Whittier, CA 90602, United States
}

\maketitle

\begin{abstract}
Based on transformed $\ell_1$ regularization, transformed total variation (TTV) has robust image recovery that is competitive with other nonconvex total variation (TV) regularizers, such as TV$^p$, $0<p<1$. Inspired by its performance, we propose a TTV-regularized Mumford--Shah model with fuzzy membership function for image segmentation. To solve it, we design an alternating direction method of multipliers (ADMM) algorithm that utilizes the transformed $\ell_1$ proximal operator. Numerical experiments demonstrate that using TTV is more effective than classical TV and other nonconvex TV variants in image segmentation.
\end{abstract}


\section{Introduction}
\label{sec:intro}

Image segmentation partitions an image into disjoint regions, where each region shares similar characteristics such as intensity, color, and texture. Penalizing the total length of the edges/boundaries of objects in an image, one fundamental model in image segmentation was proposed by Mumford and Shah \cite{mumford1989optimal}. Given a bounded, open set $\Omega \subset \R^2$ with Lipschitz boundary and an observed image $f:\Omega \to \R$, the Mumford--Shah (MS) functional to be minimized for segmentation is \begin{align*}
    \min_{u, C} \lambda \int_{\Omega} |f-u|^2 \ dx +\mu \cdot \text{ length } (C) + \int_{\Omega\backslash C} |\nabla u|^2 \ dx  
\end{align*} where $\mu$ and $\lambda$ are positive parameters, $C \subset \Omega$ is a compact curve representing the boundaries between disparate objects, and $u:\Omega \to \R$ is a piecewise-smooth function on $\Omega \backslash C$. While the MS model is robust to noise, it can be difficult to solve numerically because the unknown set of edges needs to be discretized~\cite{pock2009algorithm}.

Simplifying the MS model, Chan and Vese \cite{chan2001active} developed a model that approximates a two-region, or two-phase, image $f$ with a piecewise-constant function $u$. They model the region boundary $C$ with a Lipschitz continuous level-set function \cite{osher1988fronts}. A multiphase model \cite{vese2002multiphase} was later formulated using multiple level-set functions, but it can only be applied to images with exactly $2^k$ number of regions for $k \in \mathbb{N}$. Another method \cite{li2010multiphase} to describe the different regions of an image is by a fuzzy membership function $U= (u_1,\dots, u_N)$,  where 
\begin{equation}\label{fuzzy}
    \sum_{i=1}^N u_i(x,y) = 1, \quad 0\leq u_i(x,y) \leq 1 \text{ for } i= 1,\dots, N,
\end{equation}
at each pixel $(x,y) \in \Omega$. Under this framework, each image pixel can belong to multiple regions simultaneously with probability in $[0,1]$. As a result, the general $N$-phase segmentation problem \cite{li2010multiphase} is proposed:
\begin{align}\label{eq:fuzzy_region}
\min_{U, c} \sum_{i=1}^N \left\{ \int_{\Omega} (f-c_i)^2 u_i \; dx +\lambda \int_{\Omega} |\nabla u_i| \ dx\right\},
\end{align}
where $c = (c_1, c_2, \ldots, c_N) \in \mathbb{R}^N$. Some variants of \eqref{eq:fuzzy_region} have been developed for different cases. For example, \eqref{eq:fuzzy_region} with $\ell_1$ fidelity instead of $\ell_2$ fidelity was proposed to perform image segmentation under impulsive noise \cite{li2016multiphase}. To deal with image inhomogeneity, \eqref{eq:fuzzy_region} was revised to account for a multiplicative bias field \cite{li2010variational}. 

Since the total variation (TV) regularization $\int_{\Omega} |\nabla u| \,dx$ applies the $\ell_1$ regularization on the image gradient, noise is suppressed while edges are preserved. However, TV can result in blocky artifacts and staircase effects along the edges \cite{condat2017discrete}. Because TV regularization is a convex proxy of the $\|\nabla u\|_0$, which exactly counts the number of jump
discontinuities representing the image edges, nonconvex alternatives were investigated, showcasing more accurate results. Based on the $\ell_p,\; 0 < p < 1$, quasinorm that outperforms $\ell_1$ regularization in compressed sensing problems \cite{chartrand2009fast}, the TV$^p$ regularization found success in image restoration \cite{ hintermuller2013nonconvex}, motivating the TV$^p$ variant \cite{li2020tv} of \eqref{eq:fuzzy_region}. However, $\ell_1-\ell_2$ regularization \cite{yin2015minimization} outperformed the $\ell_p$ regularization in compressed sensing experiments, thereby motivating the development of the weighted anisotropic--isotropic TV (AITV) \cite{lou2015weighted}. An AITV variant \cite{bui2021weighted} of \eqref{eq:fuzzy_region} was later formulated.

In recent years, the transformed $\ell_1$ (TL1) regularization has been effective at robustly finding sparse solutions in signal recovery~\cite{zhang2017minimization, zhang2018minimization} and deep learning \cite{bui2021improving}. The TL1 regularization is given by
\begin{align}
    \|x\|_{\text{TL1}(a)} = \sum_{i=1}^n \rho_a(x_i), \; x \in \mathbb{R}^n,
\end{align}
where$$\rho_a(t) = \frac{(a+1)|t|}{a + |t|}, \; t\in \R, $$  and $a \in (0,\infty)$ determines the sparsity of the solution. Transformed TV (TTV) \cite{huo2022stable} extends TL1 onto the image gradient, serving as an effective regularization that can outperform TV and AITV in image recovery.  

In this paper, we propose an image segmentation model that combines fuzzy membership function and the TTV regularization. To effectively solve the model, we propose an alternating direction method of multipliers (ADMM) algorithm \cite{boyd2011distributed} utilizing the closed-form TL1 proximal operator \cite{zhang2017minimization}. Our numerical results show that the proposed model and algorithm are effective in obtaining accurate segmentation results.

\section{Proposed Approach}
For simplicity, we will use discrete notation, i.e., matrices and vectors, throughout the rest of the paper. We represent an image $f$ as an $m \times n$ matrix in the Euclidean space $X := \mathbb{R}^{m\times n}$, equipped with the usual inner product denoted by $\langle \cdot , \cdot \rangle$. 
To discretize the image gradient, we introduce the space $Y:= X\times X$ with the inner product by $\langle p,q\rangle = \langle p_1, q_1 \rangle + \langle p_2,q_2\rangle$ where $p=(p_1,p_2),\; q=(q_1,q_2) \in Y$. The discrete gradient operator $\nabla: X \rightarrow Y$ is given by $(\nabla u)_{i,j} = \left((\nabla_x u)_{i,j},
    (\nabla_y u)_{i,j} \right)$,
where $\nabla_x, \nabla_y$ are the horizontal and vertical difference operators.

The discrete version of \eqref{eq:fuzzy_region} is formulated by
\begin{align*}
    \min_{U \in S,c \in \mathbb{R}^N}  \sum_{k=1}^N \langle (f-c_k)^2, u_k\rangle + \lambda \|\nabla u_k\|_{2,1},
\end{align*}
where
\begin{align*}
    \mathcal{S} = \left\{U \in X^N: \sum_{i=1}^m \sum_{j=1}^n u_{i,j} = 1, u_{i,j} \geq 0\quad \forall i,j\right\}
\end{align*}
and $\|p\|_{2,1} = \sum_{i=1}^m \sum_{j=1}^n \sqrt{(p_1)_{i,j}^2 + (p_2)_{i,j}^2}$ for $p = (p_1,p_2) \in Y$. However, being based on the $\ell_1$ norm, TV may not be effective in edge preservation and noise suppression as its nonconvex counterparts such as TTV.  Therefore, we propose the following TTV-regularized segmentation model:
\begin{align}\label{eq:TTV_segment}
    \min_{U \in \mathcal{S}, c \in \mathbb{R}^N} \sum_{k=1}^N \langle (f-c_k)^2, u_k\rangle  + \lambda \|\nabla u_k\|_{\text{TL1}(a)},
\end{align}
where 
\begin{align}
 \|\nabla u\|_{\text{TL1}(a)} = \sum_{i=1}^m \sum_{j=1}^n \left(\rho_a(\nabla_x u_{i,j}) + \rho_a(\nabla_y u_{i,j})\right).
\end{align}
    To solve \eqref{eq:TTV_segment}, we develop an ADMM
    algorithm. We introduce auxiliary variables $V = (v_1, \ldots, v_N)$ and $D = (d_1, \ldots, d_N)$ such that $u_k = v_k$ and $d_k = \nabla v_k$ for $k=1, \ldots, N$. The augmented Lagrangian is the following:
\begin{align*}
    \mathcal{L}(&U,D, V,p,q, c)= \sum_{k=1}^N \left[\langle (f-c_k)^2, u_k\rangle+ \lambda \|d_k\|_{\text{TL}1(a)} \right]\\&+\chi_{\mathcal{S}}(U) + \langle p, U-V \rangle + \frac{\beta_1}{2} \|U-V\|_2^2\\ &+ \sum_{k=1}^N \left(\langle q_k, \nabla v_k - d_k \rangle + \frac{\beta_2}{2}\| \nabla v_k - d_k \|_2^2\right),
\end{align*}
where $\chi_S$ is an indicator function for the simplex $S$, $p$ and $q_k = ((q_x)_k, (q_y)_k), k=1, \ldots,N,$ are Lagrange multipliers, and $\beta_1, \beta_2 >0$ are penalty parameters. For each iteration $t$, the ADMM algorithm is as follows:
\begin{align*}
    U^{t+1} &\in \argmin_{U} \mathcal{L}(U,D^t, V^t, p^t, q^t, c^t),\\ 
    D^{t+1} &\in \argmin_{D} \mathcal{L}(U^{t+1},D, V^t, p^t, q^t, c^t),\\
    V^{t+1} &\in \argmin_{v} \mathcal{L}(U^{t+1},D^{t+1}, V, p^t, q^t, c^t),\\
    p^{t+1} &= p^t + \beta_1(U^{t+1} - V^{t+1}) ,\\
    q_k^{t+1} &= q_k^t +\beta_2(\nabla v_k^{t+1} - d_k^{t+1}),\; k=1,\ldots, N,\\
    c_k^{t+1} &\in \argmin_{c_k} \sum_{j=1}^N (f-c_k)^2 u_k^{t+1},\; k=1, \ldots, N.
\end{align*}

Now we describe how to solve each subproblem. With $F^t = \left(( f-c_1^t)^2, (f-c_2^t)^2, \ldots, (f-c_N^t)^2 \right)$,
the $U$-subproblem can be rewritten as
\begin{align*}
    U^{t+1} &\in  \argmin_U \langle F^t + p^t, U \rangle + \frac{\beta_1}{2} \|U - V^t\|_2^2 + \chi_{\mathcal{S}}(U)\\ &= \argmin_{U \in \mathcal{S}} \frac{\beta_1}{2} \left\|  U-\left(V^t - \frac{F^t+p^t}{\beta_1} \right)\right\|_2^2\\
    &= \text{proj}_S \left(V^t - \frac{F^t+p^t}{\beta_1}\right),
\end{align*}
where $\text{proj}_S$ is the projection onto a simplex. The simplex projection algorithm is described in \cite{chen2011projection}. Each $k$th component of the $D$-subproblem can be solved individually as follows:
\begin{align}\label{eq:u_sub1}
    d_k^{t+1} &\in \argmin_{d_k} \lambda \|d_k\|_{\text{TL1}(a)} + \frac{\beta_2}{2} \left\|d_k -\left(\nabla v_k^t + \frac{q_k^t}{\beta_2} \right)\right\|_2^2.
\end{align}
Reducing \eqref{eq:u_sub1} at each pixel $(i,j)$, we have
\begin{gather}
\begin{aligned}\label{eq:d_update}
    (d_k^{t+1})_{i,j}= \Bigg(&\text{prox}_{\frac{\lambda}{\beta_2} \|\cdot\|_{\text{TL1}(a)}} \left(\nabla_x (v_k)_{i,j}^t + \frac{((q_x)_k)_{i,j}^t}{\beta_2} \right),\\
    &\text{prox}_{\frac{\lambda}{\beta_2} \|\cdot\|_{\text{TL1}(a)}} \left(\nabla_y (v_k)_{i,j}^t + \frac{((q_y)_k)_{i,j}^t}{\beta_2} \right)\Bigg),
\end{aligned}
\end{gather}
where 
\begin{align}\label{eq:tl1_prox} \text{prox}_{\lambda \|\cdot \|_{\text{TL1}(a)}} (x) = \argmin_y \|y\|_{\text{TL1}(a)} + \frac{1}{2 \lambda}\|y-x\|_2^2
\end{align}
is the proximal operator for TL1. The TL1 proximal operator has a closed-form solution provided in the lemma below.
\begin{lemma}[\cite{zhang2017minimization}]
Given $x \in \mathbb{R}^n$, the optimal solution to \eqref{eq:tl1_prox} is 
\end{lemma}
\begin{align*}\text{prox}_{\lambda \| \cdot \|_{\text{TL1}(a)}}(x) = \left(\mathcal{T}_{a, \lambda}(x_1), \ldots, \mathcal{T}_{a, \lambda}(x_n) \right),
\end{align*} 
with
\begin{equation*}
    \mathcal{T}_{a, \lambda}(t) =\begin{cases}
    0 &\text{ if } |t| \leq \tau(a, \lambda), \\
    g_{a,\lambda}(t) &\text{ if } |t| > \tau(a, \lambda)
    \end{cases}
\end{equation*}
where 
\begin{equation*}
    g_{a,\lambda}(t) = \text{sign}(t) \left(\frac{2}{3}(a+|t|) \cos \left( \frac{\phi_{a, \lambda}(t)}{3} \right) - \frac{2a}{3} + \frac{|t|}{3} \right),
\end{equation*}
\begin{equation*}
    \phi_{a, \lambda}(t) = \arccos \left(1-\frac{27\lambda a (a+1)}{2(a+|t|)^3} \right),
\end{equation*}
and
\begin{equation*}
    \tau(a,\lambda) = \begin{cases}
    \sqrt{2 \lambda (a+1)} - \frac{a}{2} & \text{ if } \lambda > \frac{a^2}{2(a+1)}, \\
    \lambda \frac{a+1}{a} & \text{ if } \lambda \leq \frac{a^2}{2(a+1)}.
    \end{cases}
\end{equation*}
The $V$-subproblem is separable with respect to each $v_k$, so it can be simplified as
\begin{align*}
    v_k^{t+1} \in &\argmin_{v_k} \langle p_k^t, u_k^{t+1}-v_k\rangle + \frac{\beta_1}{2} \|u_k^{t+1}-v_k\|_2^2\\ &+ \langle q_k^t, \nabla v_k - d_k^{t+1} \rangle + \frac{\beta_2}{2}\| \nabla v_k - d_k^{t+1} \|_2^2,
\end{align*}
which is equivalent to the first-order optimality condition
\begin{align*}
    (\beta_1 I - \beta_2 \Delta) v_k = p_k^t + \beta_1 u_k^{t+1} -\nabla^{\top}(q_k^t -\beta_2 d_k^{t+1}).
\end{align*}
Assuming periodic boundary condition, we can solve this via Fourier Transform $\mathcal{F}$ \cite{wang2008new}. Hence, we have
\begin{align*}
    v_k^{t+1} = \mathcal{F}^{-1} \left( \frac{ \mathcal{F}(p_k^t + \beta_1 u_k^{t+1} - \nabla^{\top}(q_k^t - \beta_2 d_k^{t+1}))}{\beta_1 I - \beta_2 \mathcal{F}(\Delta)} \right).
\end{align*}
Lastly, solving the $c_k$-subproblem is equivalent to
\begin{align*}
    c_k^{t+1} =  \frac{\displaystyle\sum_{i,j} f_{i,j} (u_k^{t+1})_{i,j}}{\displaystyle \sum_{i,j} (u_k^{t+1})_{i,j}}.
\end{align*}

The overall ADMM algorithm to solve \eqref{eq:TTV_segment} is summarized in Algorithm \ref{alg:admm}.
\begin{algorithm}[h!!!]
\caption{ADMM for \eqref{eq:TTV_segment}}
\label{alg:admm}
\scriptsize
\begin{algorithmic}[1]
\REQUIRE Image $f$; regularization parameter $\lambda$; sparsity parameter $a$; penalty parameter $\beta_1, \beta_2$.\\
    \STATE Initialize $V^0, p^0, q^0, c^0$.\\
    \STATE Set $t=0$.\\
   \WHILE{$\frac{\|U^t-U^{t-1}\|_F}{\|U^t\|_F} > \epsilon$}
   \STATE Compute $U^{t+1} = \text{proj}_S \left(V^t - \frac{F^t+p^t}{\beta_1}\right)$. See \cite{chen2011projection} for projection algorithm.
   \STATE Compute $D^{t+1}$ according to \eqref{eq:d_update}.
   \STATE Compute $v_k^{t+1} = \mathcal{F}^{-1} \left( \frac{ \mathcal{F}(p_k^t + \beta_1 u_k^{t+1} -\nabla^{\top}(q_k^t - \beta_2 d_k^{t+1}))}{\beta_1 I - \beta_2 \mathcal{F}(\Delta)} \right)$ for each\\ $k=1,\ldots, N$.
   \STATE Compute $p^{t+1} = p^t + \beta_1(U^{t+1} - V^{t+1})$
   \STATE  Compute $q_k^{t+1} = q_k^t +\beta_2(\nabla v_k^{t+1} - d_k^{t+1})$ for each $k=1, \ldots, N$.
    \STATE Compute $c_k^{t+1} =  \frac{\displaystyle\sum_{i,j} f_{i,j} (u_k^{t+1})_{i,j}}{\displaystyle \sum_{i,j} (u_k^{t+1})_{i,j}}$ for each $k=1, \ldots, N$.
    \STATE $t \coloneqq t+1$.
   \ENDWHILE
   \RETURN $U^t, C^t$.\\
\end{algorithmic}
\end{algorithm}

\begin{figure}
\centering
               \begin{subfigure}[b]{0.10\textwidth}
         \centering
         \includegraphics[width=\textwidth]{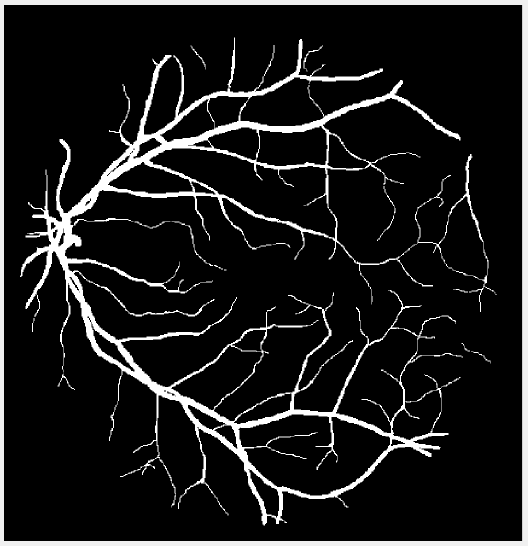}
         \caption{Vessel 1}
         \label{fig:vessel1}
     \end{subfigure}
     \begin{subfigure}[b]{0.10\textwidth}
         \centering
         \includegraphics[width=\textwidth]{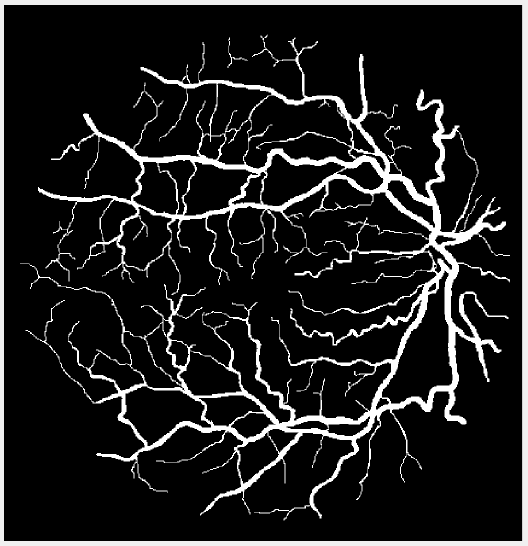}
         \caption{Vessel 2}
         \label{fig:vessel2}
     \end{subfigure}
        \begin{subfigure}[b]{0.10\textwidth}
         \centering
         \includegraphics[width=\textwidth]{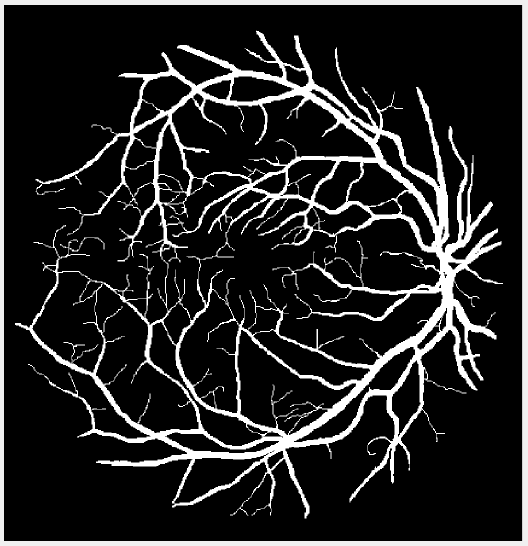}
         \caption{Vessel 3}
         \label{fig:vessel3}
     \end{subfigure}\\
     \begin{subfigure}[b]{0.10\textwidth}
         \centering
         \includegraphics[scale=0.15]{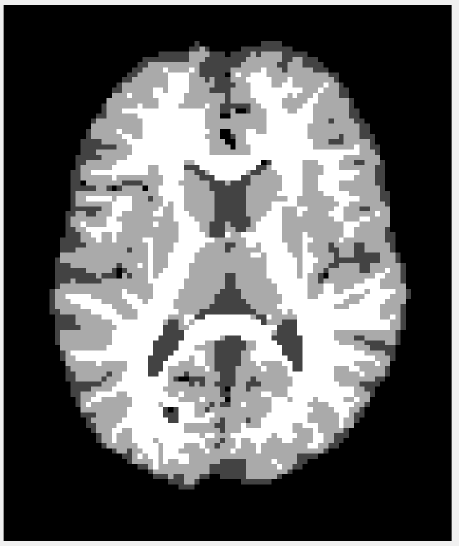}
         \caption{Brain 1}
         \label{fig:brain1}
     \end{subfigure}
          \begin{subfigure}[b]{0.10\textwidth}
         \centering
         \includegraphics[scale=0.15]{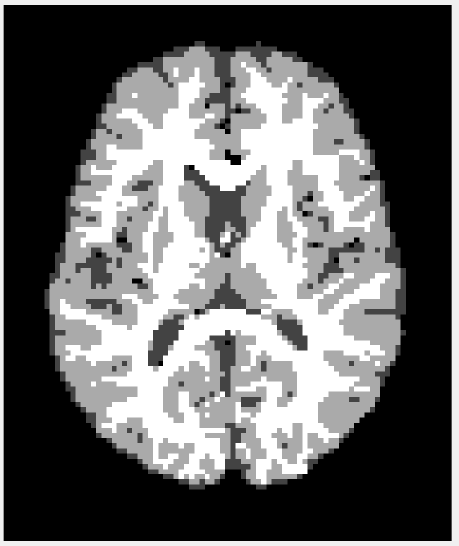}
         \caption{Brain 2}
         \label{fig:brain2}
     \end{subfigure}
          \begin{subfigure}[b]{0.10\textwidth}
         \centering
         \includegraphics[scale=0.15]{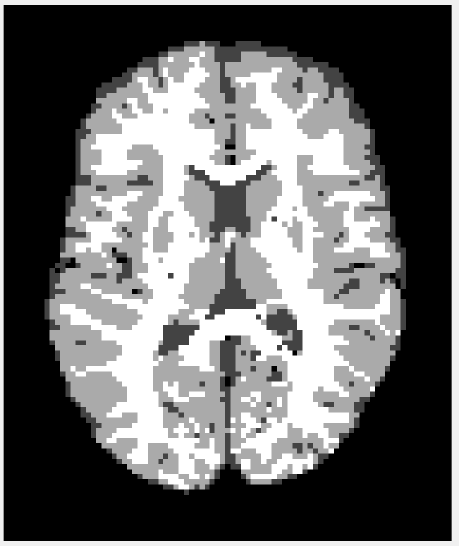}
         \caption{Brain 3}
         \label{fig:brain3}
     \end{subfigure}
               \begin{subfigure}[b]{0.10\textwidth}
         \centering
         \includegraphics[scale=0.15]{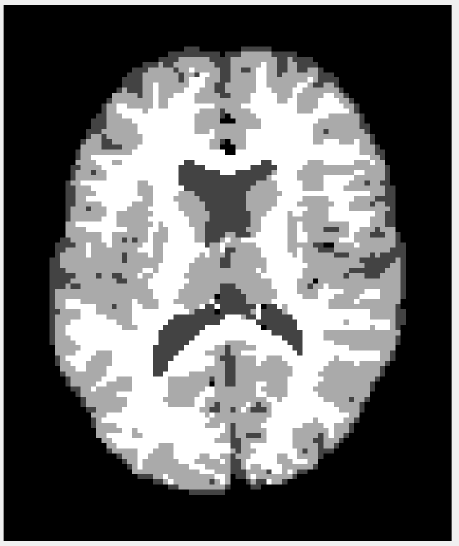}
         \caption{Brain 4}
         \label{fig:brain4}
     \end{subfigure}
        \caption{Original images for testing. (a)-(c) Retinal vessel images from the DRIVE dataset \cite{staal2004ridge}. Image size is $584 \times 565$ with pixel intensities 191 (vessel) and 104 (background). (d)-(g) Brain images from the BrainWeb dataset \cite{aubert2006new}. Image size is $104 \times 87$ with pixel intensities 10 (background), 48 (cerebrospinal fluid), 106 (grey matter), and 154 (white matter). }
        \label{fig:experiment_image}
\end{figure}
\section{Experimental Results}
We compare the performance of the proposed TTV-regularized image segmentation model with its counterparts regularized by (isotropic) TV \cite{li2010variational}, TV$^p$ \cite{li2020tv}, and AITV \cite{bui2021weighted}. The algorithm we use for TV is similar to Algorithm \ref{alg:admm} in that we use the $\ell_{2,1}$ proximal operator in \eqref{eq:d_update}. For TV$^p$, we use the ADMM algorithm following \cite{li2020tv} but without the bias term for fair comparison and set $p=1/3$ as suggested. For AITV, we use the difference-of-convex algorithm (DCA) \cite{tao1997convex} designed in \cite{bui2021weighted} and set $\alpha = 0.5$ as suggested. The image segmentation models are applied to the images shown in Figure \ref{fig:experiment_image}. For Figures \ref{fig:vessel1}-\ref{fig:vessel3}, we perform binary segmentation to identify the retina vessels, while for Figures \ref{fig:brain1}-\ref{fig:brain4}, we perform multiphase segmentation ($N=4$) to identify the cerebrospinal fluid (CSF), grey matter (GM), and white matter (WM) separate from the background. We evaluate the segmentation performance by two metrics: DICE index \cite{dice1945measures} and Jaccard similarity index \cite{jaccard1901distribution}. The parameters for each segmentation method are carefully tuned so that we obtain the best DICE indices. Specifically, for Algorithm \ref{alg:admm} that solves \eqref{eq:TTV_segment}, we set $\beta_1=\beta_2 = 0.25$ and find the optimal parameter $\lambda$ in the range $[0.0025, 0.05]$ for both binary and multiphase segmentation. For binary segmentation, we select the best value for $a \in \{5, 10, 100\}$ while for multiphase segmentation, we select for $a \in \{1, 5, 10\}$. Algorithm \ref{alg:admm} is initialized with the results of fuzzy $c$-means clustering \cite{bezdek1984fcm} and it terminates either when $\frac{\|U^t-U^{t-1}\|_F}{\|U^t\|_F}< 10^{-4}$ or after 200 iterations. The experiments are
performed in MATLAB R2022b on a Dell laptop with a
1.80 GHz Intel Core i7-8565U processor and 16.0 GB of
RAM. The code for Algorithm \ref{alg:admm} is available at \url{https://github.com/JimTheBarbarian/Official-TTV-Segmentation}.

\begin{figure*}
\centering
               \begin{subfigure}[b]{\textwidth}
         \centering
         \includegraphics[scale=0.45]{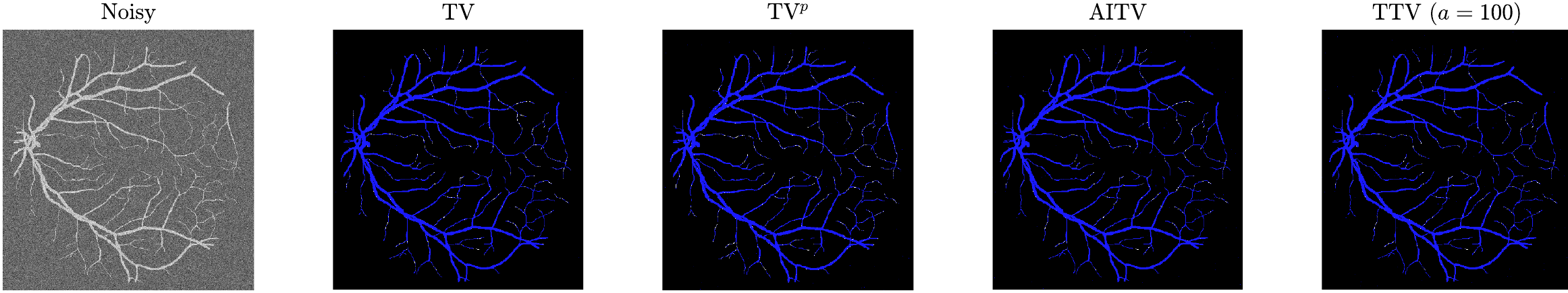}
         \caption{Vessel 1}
         \label{fig:segment_vessel1}
     \end{subfigure}
     \begin{subfigure}[b]{\textwidth}
         \centering
         \includegraphics[scale=0.45]{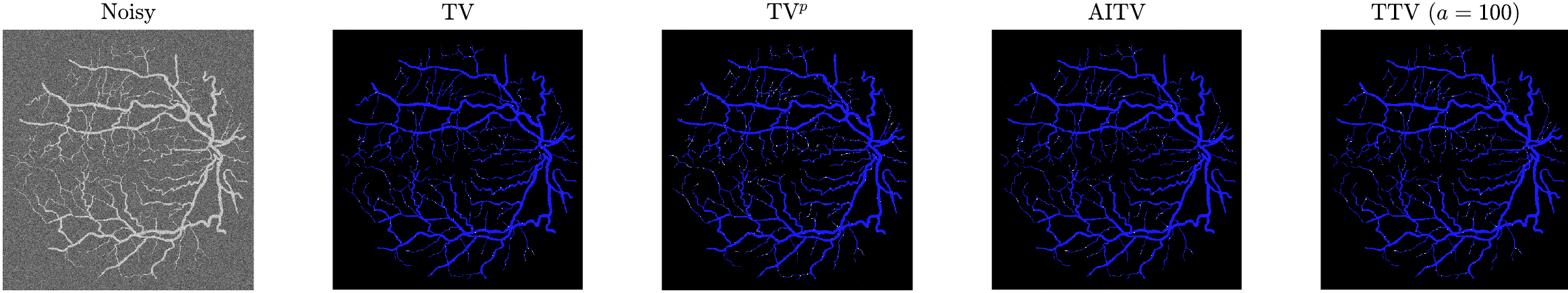}
         \caption{Vessel 2}
         \label{fig:segment_vessel2}
     \end{subfigure}
        \caption{Segmentation results of Figures \ref{fig:vessel1}-\ref{fig:vessel2} (after normalization) corrupted by Gaussian noise of mean 0 and variance 0.01. }
        \label{fig:retina_segment}
\end{figure*}

\begin{figure}
    \centering
    \includegraphics[scale=0.5]{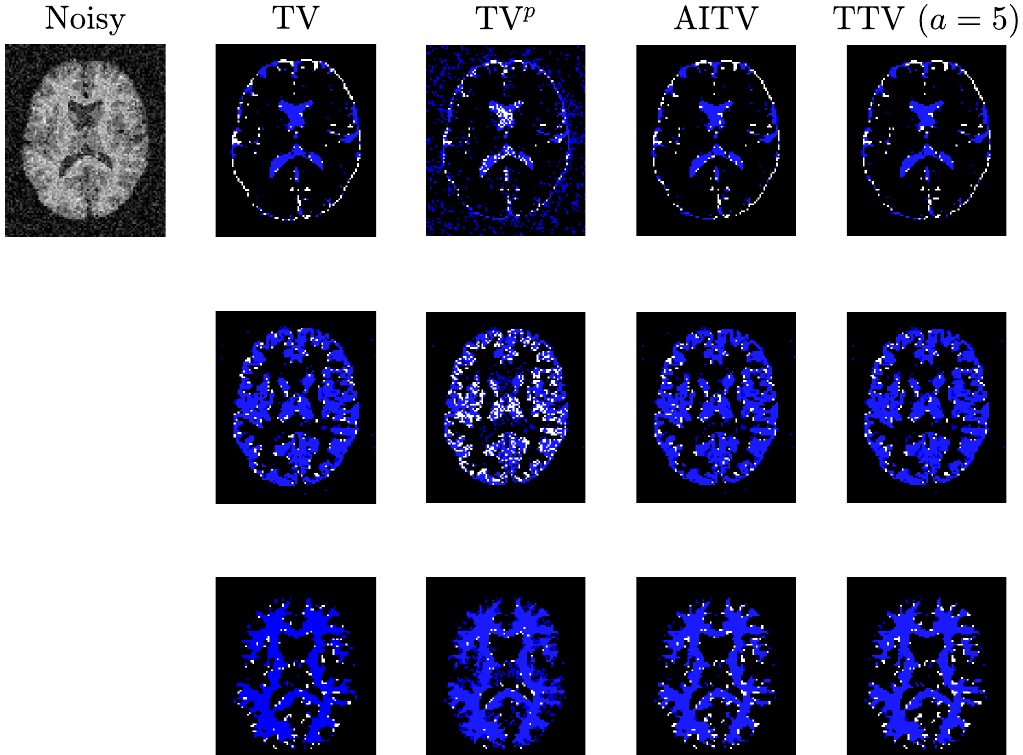}
    \caption{Segmentation results of Figure \ref{fig:brain4} (after normalization) corrupted by Gaussian noise of mean 0 and variance 0.04. First row is CSF, second row is GM, and third row is WM. }
    \label{fig:brain_segment4}
\end{figure}

Before applying the segmentation algorithms, the images in Figure \ref{fig:experiment_image} are normalized to $[0,1]$ followed by Gaussian noise corruption. The retina vessel images are corrupted with Gaussian noise of mean 0 and variance 0.01. Table \ref{tab:binary_segment} reports the performances and times of the segmentation methods on the retina vessel images while Figure \ref{fig:retina_segment} shows some of their results. TTV $(a=10, 100)$ has the highest DICE and Jaccard similarity indices across the three images although requiring about 80 seconds to complete, thereby being slower than TV and TV$^p$.  The brain images are corrupted with Gaussian noise of mean 0 and variance 0.04. Table \ref{tab:multi_segment} reports the performances and times of the multiphase segmentation, while Figure \ref{fig:brain_segment4} shows the segmentation results of Figure \ref{fig:brain4}. By its DICE and Jaccard similarity indices, TTV $(a=1)$ is best at segmenting CSF across the four images while TTV $(a=5,10)$ remains competitive against AITV in segmenting GM and WM. On average, TTV $(a=1,5)$ is among the top two best-performing methods. Although it can be outperformed by AITV, it is at least three times faster. In Figure \ref{fig:brain_segment4}, we see that TTV $(a=5)$ is most effective in segmenting CSF, especially compared to TV and TV$^p$. Moreover, comparable to TV and AITV, it is able to identify most of the GM and WM regions. Overall, using TTV, the proposed method is able to effectively identify narrow, thin regions such as the retina vessels and CSF.

\section{Conclusion}
We designed a segmentation model that combines fuzzy membership functions and TTV and developed an ADMM algorithm to solve it.  Our experiments demonstrated that TTV is an effective regularizer for image segmentation, especially when segmenting narrow regions where edge preservation is important. For future directions, we will extend our proposed model to color images.

\begin{table}[t!]
\centering
\scriptsize
\caption{Accuracy and times (sec.) of image segmentation methods on Figures \ref{fig:vessel1}-\ref{fig:vessel3} (after normalization) corrupted by Gaussian noise of mean 0 and variance 0.01. \textbf{Bold} indicates best value.}
\label{tab:binary_segment}
\begin{subtable}[h]{0.5\textwidth}
\centering
\caption{DICE and Jaccard similarity indices.}
\begin{tabular}{|l|c|c|c|c|c|c|}
\cline{2-7}
\multicolumn{1}{c}{} &\multicolumn{2}{|c|}{Vessel 1}     & \multicolumn{2}{|c|}{Vessel 2}  & \multicolumn{2}{|c|}{Vessel 3} \\ \cline{2-7}
\multicolumn{1}{c|}{} & DICE & Jaccard & DICE & Jaccard & DICE & Jaccard \\ \hline
TV & 0.9777 &	0.9564	&0.9772 &	0.9555& 0.9812	& 0.9630
\\ \hline
TV$^p$ & 0.9640 &	0.9304	& 0.9653	& 0.9330 & 0.9708&	0.9433
 \\ \hline
AITV & 0.9786	& 0.9581	& 0.9777	& 0.9564 & 0.9806	& 0.9619
\\ \hline
TTV $(a=5)$ & 0.9788	&0.9585	& 0.9773	&0.9556 &0.9806	&0.9619
 \\ \hline
TTV $(a=10)$ & 0.9799	& 0.9605&	0.9794&	0.9596 &\textbf{0.9821}	&\textbf{0.9648} \\ \hline
 TTV $(a=100)$ & \textbf{0.9801}	& \textbf{0.9610} &	\textbf{0.9796}	& \textbf{0.9599} & 0.9820 &	0.9647
 \\ \hline
\end{tabular}
\end{subtable}
\\
\vspace{0.05cm}
\begin{subtable}[h]{0.5\textwidth}
\caption{Computational time (sec.).}
\centering
\begin{tabular}{|c|c|c|c||c|} \cline{2-5}
    \multicolumn{1}{c|}{} & Vessel 1  & Vessel 2 & Vessel 3 & Avg. \\  \hline
     TV & 47.57	& \textbf{52.19}	& \textbf{45.64}	& \textbf{48.47} \\ \hline
     TV$^p$ & \textbf{38.33} &	70.25	&65.13 &	57.90\\ \hline
AITV	& 250.28	& 251.01	& 172.80	& 224.70\\ \hline
TTV ($a=5$) & 92.40 &	73.93	&77.26	& 81.20\\ \hline
TTV$(a=10)$	& 87.68 &	79.66&	74.03&	80.46\\ \hline 
TTV$(a=100)$	& 88.82	& 73.27	& 72.86 &	78.32 \\ \hline
\end{tabular}
\end{subtable}
\end{table}

\begin{table*}
\centering
\scriptsize
\caption{Accuracy and times (sec.) of image segmentation methods on Figures \ref{fig:brain1}-\ref{fig:brain4} (after normalization) corrupted by Gaussian noise of mean 0 and variance 0.04. For DICE and Jaccard, the average for each image is computed across the three regions: cerebrospinal fluid (CSF), grey matter (GM), and white matter (WM). \textbf{Bold} indicates best value.}
\label{tab:multi_segment}
\begin{subtable}[h]{\textwidth}
\centering
\caption{DICE index.}
\resizebox{\textwidth}{!}{\begin{tabular}{|l|c|c|c|c|c|c|c|c|c|c|c|c|c|c|c|c|}
\cline{2-17}
\multicolumn{1}{c}{} &\multicolumn{4}{|c|}{Brain 1}     & \multicolumn{4}{|c|}{Brain 2}  & \multicolumn{4}{|c|}{Brain 3} & \multicolumn{4}{|c|}{Brain 4} \\ \cline{2-17}
\multicolumn{1}{c|}{} & CSF & GM & WM & Avg. & CSF & GM & WM & Avg. & CSF & GM & WM & Avg. & CSF & GM & WM & Avg. \\ \hline
        TV & 0.7650 & 0.8953 & 0.9110 & 0.8571 & 0.6843 & 0.8824 & 0.9056 & 0.8241 & 0.6595 & 0.8900 & \textbf{0.9090} & 0.8195 & 0.7711 & 0.8992 & 0.9213 & 0.8639 \\ \hline
        TV$^p$ & 0.7724 & 0.8489 & 0.8614 & 0.8276 & 0.6097 & 0.7710 & 0.7954 & 0.7254 & 0.6692 & 0.8632 & 0.8882 & 0.8069 & 0.5038 & 0.7617 & 0.8552 & 0.7069 \\ \hline
        AITV & 0.7711 & 0.8946 & \textbf{0.9114} & \textbf{0.8590} & 0.6979 & \textbf{0.8883} & \textbf{0.9079} & \textbf{0.8314} & 0.7038 & 0.8882 & 0.9036 & 0.8319 & 0.7683 & 0.8934 & 0.9146 & 0.8588 \\ \hline
        TTV$(a=1)$ & \textbf{0.7810} & 0.8937 & 0.9023 & \textbf{0.8590} & \textbf{0.7170} & 0.8787 & 0.8902 & 0.8286 & \textbf{0.7203} & 0.8865 & 0.8953 & \textbf{0.8340} & \textbf{0.7910} & 0.8859 & 0.9042 & 0.8604 \\ \hline
        TTV$(a=5)$ & 0.7643 & 0.8961 & 0.9093 & 0.8566 & 0.7054 & 0.8855 & 0.9029 & 0.8313 & 0.6827 & 0.8921 & 0.9087 & 0.8278 & 0.7827 & 0.8974 & 0.9177 & \textbf{0.8659} \\ \hline
        TTV$(a=10)$ & 0.7527 & \textbf{0.8984} & 0.9107 & 0.8539 & 0.6987 & 0.8767 & 0.8978 & 0.8244 & 0.6755 & \textbf{0.8934} & \textbf{0.9090} & 0.8260 & 0.7620 & \textbf{0.8997} & \textbf{0.9215} & 0.8611 \\ \hline

\end{tabular}}
\end{subtable}
\\
\vspace{0.05cm}
\begin{subtable}[h]{\textwidth}
\centering
\caption{Jaccard similarity index.}
\resizebox{\textwidth}{!}{\begin{tabular}{|l|c|c|c|c|c|c|c|c|c|c|c|c|c|c|c|c|}
\cline{2-17}
\multicolumn{1}{c}{} &\multicolumn{4}{|c|}{Brain 1}     & \multicolumn{4}{|c|}{Brain 2}  & \multicolumn{4}{|c|}{Brain 3} & \multicolumn{4}{|c|}{Brain 4} \\ \cline{2-17}
\multicolumn{1}{c|}{} & CSF & GM & WM & Avg. & CSF & GM & WM & Avg. & CSF & GM & WM & Avg. & CSF & GM & WM & Avg. \\ \hline
        TV & 0.6195 & 0.8104 & 0.8366 & 0.7555 & 0.5201 & 0.7924 & 0.8274 & 0.7133 & 0.4920 & 0.8017 & \textbf{0.8332} & 0.7090 & 0.6274 & 0.8168 & 0.8541 & 0.7661 \\ \hline
        TV$^p$ & 0.6292 & 0.7374 & 0.7565 & 0.7077 & 0.4386 & 0.6274 & 0.6604 & 0.5755 & 0.5028 & 0.7594 & 0.7989 & 0.6870 & 0.3368 & 0.6152 & 0.7470 & 0.5663 \\ \hline
        AITV & 0.6275 & 0.8093 & \textbf{0.8372} & \textbf{0.7580} & 0.5360 & \textbf{0.7990} & \textbf{0.8313} & \textbf{0.7221} & 0.5430 & 0.7989 & 0.8241 & 0.7220 & 0.6238 & 0.8073 & 0.8427 & 0.7579 \\ \hline
        TTV$(a=1)$ & \textbf{0.6407} & 0.8078 & 0.8220 & 0.7568 & \textbf{0.5588} & 0.7836 & 0.8022 & 0.7149 & \textbf{0.5629} & 0.7961 & 0.8105 & \textbf{0.7232} & \textbf{0.6542} & 0.7952 & 0.8251 & 0.7582 \\ \hline
        TTV$(a=5)$ & 0.6185 & 0.8118 & 0.8337 & 0.7547 & 0.5449 & 0.7945 & 0.8230 & 0.7208 & 0.5182 & 0.8051 & 0.8327 & 0.7187 & 0.6429 & 0.8138 & 0.8479 & \textbf{0.7682} \\ \hline
        TTV$(a=10)$ & 0.6034 & \textbf{0.8156} & 0.8360 & 0.7517 & 0.5369 & 0.7805 & 0.8145 & 0.7106 & 0.5100 & \textbf{0.8074} & \textbf{0.8332} & 0.7169 & 0.6155 & \textbf{0.8178} & \textbf{0.8545} & 0.7626 \\ \hline

\end{tabular}}
\end{subtable}
\\
\vspace{0.05cm}
\begin{subtable}[h]{\textwidth}
\centering
    \caption{Computational time (sec.).}
    \begin{tabular}{|l|c|c|c|c|c|}
    \cline{2-6}
    \multicolumn{1}{c|}{} & Brain 1 & Brain 2 & Brain 3 & Brain 4 & Avg. \\ \hline
            TV & \textbf{3.25} & \textbf{4.24} & 3.28 & \textbf{3.27} & \textbf{3.51} \\ \hline
        TV$^p$ & 8.04 & 9.24 & 7.76 & 8.09 & 8.28 \\ \hline
        AITV & 15.23 & 16.36 & 14.20 & 14.43 & 15.06 \\ \hline
        TTV$(a=1)$ & 4.34 & 4.65 & \textbf{3.06} & 3.78 & 3.96 \\ \hline
        TTV$(a=5)$ & 4.71 & 5.92 & 4.57 & 4.90 & 5.03 \\ \hline
        TTV$(a=10)$ & 4.76 & 6.31 & 4.64 & 4.73 & 5.11 \\ \hline
    \end{tabular}
\end{subtable}
\end{table*}
\bibliographystyle{IEEEtran}
\bibliography{refs}

\vspace{12pt}

\end{document}